\documentclass[11pt,a4paper]{article}
\pdfoutput=1  
\usepackage[utf8]{inputenc}
\usepackage[hyperref]{acl2019}
\usepackage{times}
\usepackage{comment}
\usepackage{latexsym}
\usepackage{breqn}
\usepackage{adjustbox}
\usepackage{tabulary}
\usepackage{tabularx}
\usepackage{multirow}
\usepackage[dash,dot]{dashundergaps}

\usepackage{graphicx}
\usepackage{subfigure} 

\usepackage{url}

\usepackage{lingmacros}

\aclfinalcopy

\title{Constructing large scale biomedical knowledge bases from scratch with rapid annotation of interpretable patterns}

\author{
Julien Fauqueur\thanks{\hspace{0.1cm} Equal contribution. Listing order is alphabetical. 
Theodosia proposed and co-ordinated the research project, built the early prototypes and contributed the different methods for extracting and lexicalising patterns. 
Ashok provided conceptual work on the metrics for ranking simplifications and for the intrinsic evaluation, developed the simplification extraction module, ran the experiments for the automated workflow (with all the parameter variations) and performed all the extrinsic evaluations.
Julien was mainly responsible for the system architecture and workflow, the intrinsic evaluation (including interacting with the experts), handling negation and speculation and the clustering algorithm.} \\
  BenevolentAI \\
  4-8 Maple St, London \\
   W1T 5HD \\
  \texttt{julien@benevolent.ai} \\\And
  Ashok Thillaisundaram\footnotemark[1] \\
  BenevolentAI \\
  4-8 Maple St, London \\
   W1T 5HD \\
  \texttt{ashok@benevolent.ai} \\\And
  Theodosia Togia\footnotemark[1] \\
  BenevolentAI \\
  4-8 Maple St, London \\
   W1T 5HD \\
  \texttt{sia@benevolent.ai}\\
}
  
\date{May 2019}

\begin{document}

\maketitle

\begin{abstract}

Knowledge base construction is crucial for summarising, understanding and inferring relationships between biomedical entities. However, for many practical applications such as drug discovery, the scarcity of relevant facts (e.g. \textit{gene X is therapeutic target for disease Y}) severely limits a domain expert's ability to create a usable knowledge base, either directly or by training a relation extraction model. In this paper, we present a simple and effective method of extracting new facts with a pre-specified binary relationship type from the biomedical literature, without requiring any training data or hand-crafted rules. Our system discovers, ranks and presents the most salient patterns to domain experts in an interpretable form. By marking patterns as compatible with the desired relationship type, experts indirectly batch-annotate candidate pairs whose relationship is expressed with such patterns in the literature. Even with a complete absence of seed data, experts are able to discover thousands of high-quality pairs with the desired relationship within minutes. When a small number of relevant pairs do exist - even when their relationship is more general (e.g. \textit{gene X is biologically associated with disease Y}) than the relationship of interest - our system leverages them in order to i) learn a better ranking of the patterns to be annotated or ii) generate weakly labelled pairs in a fully automated manner. We evaluate our method both intrinsically and via a downstream knowledge base completion task, and show that it is an effective way of constructing knowledge bases when few or no relevant facts are already available.

\end{abstract}

\section{Introduction}

In many important biomedical applications, experts seek to extract facts that are often complex and tied to particular tasks, hence data that are truly fit for purpose are scarce or simply non-existent. Even when only binary relations are sought, useful facts tend to be more specific (e.g. \textit{mutation of gene X has a causal effect on disease Y in an animal model}) than associations typically found in widely available knowledge bases. Extracting facts with a pre-specified relationship type from the literature in the absence of training data often relies on handcrafted rules, which are laborious, ad-hoc and hardly reusable for other types of relations. Recent attempts to create relational data from scratch by denoising the output of multiple hand-written rules \cite{dataprog} or by augmenting existing data through the induction of new black-box heuristics \cite{snuba} are still dependent on ad-hoc human effort or pre-existing data. Our approach involves discovering and recommending, rather than prescribing, rules. Importantly, our rules are presented as text-like patterns whose meaning is transparent to human annotators, enabling integration of an automatic data generation (or augmentation) system with a domain expert feedback loop.

In this work, we make the following contributions:

\begin{itemize}
    \vspace{-0.1cm}\item We propose a number of methods for extracting patterns from a sentence in which two eligible entities co-occur; different types of patterns have different trade-offs between expressive power and coverage.
    \vspace{-0.15cm}\item We propose a simple method for presenting patterns in a readable way, enabling faster, more reliable human annotation
    \vspace{-0.15cm}\item For cases where a small number of seed pairs are already available, we propose a method which utilises these seed pairs to rank newly discovered patterns in terms of their compatibility with the existing data. The resulting patterns can be used with or without a human in the loop.
\end{itemize}

The rest of the paper is organised as follows. Section \ref{related-work} describes some related work. Section \ref{extracting-and-lexicalising-patterns} explains the relationship between patterns and labelling rules and presents some pattern types along with techniques for rendering them interpretable. Section \ref{system-overview} provides a high-level overview of the system and covers details of our different workflows (with and without seed data; with and without human feedback). Section \ref{evaluation} explains how we measure the system's performance both intrinsically and via a downstream knowledge base completion task. In section \ref{experimental-setup}, we report the details of our main experiments while in sections \ref{top-simplifications} and \ref{further-experiments} we present some analysis along with further experiments. The paper ends with conclusions and proposals for further work in section \ref{conclusions-and-further-work}.

\section{Related work}\label{related-work}

The idea of extracting entity pairs by discovering textual patterns dates back to early work on bootstrapping for relation extraction with the DIPRE system \cite{dipre}. This system was designed to find co-occurrences of seed entity pairs of a known relationship type inside unlabelled text, then extract simple patterns (exact string matches) from these occurrences and use them to discover new entity pairs. Agichtein et al.~\shortcite{snowball} introduced a pattern evaluation methodology based on the precision of a pattern on the set of entity pairs which had already been discovered; they also used the dot product between word vectors instead of an exact string match to allow for slight variations in text. Later work \cite{Greenwood:2006,xu:bootstrap, Alfonseca:2012} has proposed more sophisticated pattern extraction methods (based on dependency graphs or kernel methods on word vectors) and different pattern evaluation frameworks (document relevance scores).

Two recent weak supervision techniques, Data Programming \cite{dataprog} and the method underlying the Snuba system \cite{snuba} have attempted to combine the results of handcrafted rules and weak base classifiers respectively. Data Programming involves modelling the accuracy of ideally uncorrelated rules devised by domain experts, then combining their output into weak labels. Although this approach does not require any seed data, it does rely on handwritten rules, which are both time consuming and ad-hoc due to the lack of a data-driven mechanism for exploring the space of possible rules. Snuba learns black-box heuristics (parameters for different classifiers) given seed pairs of the desired relationship. This method avoids the need for manually composing rules, however, the rules it learns are not interpretable, which makes the pipeline harder to combine with an active learning step. Second, the system requires gold standard pairs. In contrast, while our system can leverage gold standard annotations, if available, in order to reduce the space of discovered rules, as well as tune the ranking of newly discovered patterns, it is entirely capable of starting without any gold data if ranking is heuristics-based (e.g. prioritisation by frequency) and a human assesses the quality of the highest coverage rules suggested. Our method does not preclude use within a data programming setup as a way of discovering labelling functions or within a system like Snuba, as a way of generating seed pairs. Another body of work, distant supervision \cite{bran, nresai} has been a recent popular way to extract relationships from weak labels, but does not give the user any control on the model performance.

A well known body of work, OpenIE \cite{textrunner,reverb,ollie,stanfordOpenIE15} aims to extract patterns between entity mentions in sentences, thereby discovering new surface forms which can be clustered \cite{mohamed2011discovering,journals/pvldb/NakasholeWS12} in order to reveal new meaningful relationship types. In the biomedical domain, Percha and Altman~\shortcite{journals/bioinformatics/PerchaA18} attempt something similar by extracting and clustering dependency patterns between pairs of biomedical entities (e.g. chemical-gene, chemical-disease, gene-disease). Our work differs from these approaches in that we extract pairs for a pre-specified relationship type (either from scratch or by augmenting existing data written with specific guidelines), which is not guaranteed to correspond to a cluster of discovered surface forms.

\section{Extracting interpretable patterns}\label{extracting-and-lexicalising-patterns}

In a rule-based system, a rule, whether handwritten or discovered, can be described as a hypothetical proposition ``$if ~ P ~ then ~ Q$", where $P$ (the antecedent) is a set of conditions that may be true or false of the system's input and $Q$ (the consequent) is the system's output. For instance, a standard rule-based relation extraction system can \textbf{i)} take as input a pair of entities (e.g. \texttt{\small{TNF-GeneID:7124}} and \texttt{\small{Melanoma-MESH:D008545}}) that are mentioned in the same piece of text, \textbf{ii)} test whether certain conditions are met (e.g. presence of lexical or syntactic features) and \textbf{iii)} output a label (e.g. \texttt{\small{1: Therapeutic target, 0: Not therapeutic target}}.) 

In this work, patterns are seen as the antecedents of rules that determine which label (consequent) should be assigned to some input (e.g. candidate pair + text that mentions it.) We aim to extract patterns that are expressive enough to allow a system or a domain expert to discriminate between the different labels available for an input but also generic enough to apply to a wide range of inputs. In this work, we have made the following simplifying assumptions:
\begin{enumerate}
    \vspace{-0.1cm}\item Relationships are binary (i.e. hold between exactly two entities).
    \vspace{-0.15cm}\item A pair of entities are candidates for relation extraction if they are mentioned simultaneously in the same sentence.
    \vspace{-0.15cm}\item There is a one-to-many relationship between patterns and inputs. An input (i.e. sentence + entity pair) is described by a single pattern (although this pattern can be a boolean combination of other patterns) but one pattern can correspond to multiple inputs.
    \vspace{-0.15cm}\item We can select patterns which are expressive enough to represent the relationship, so it is possible to classify the input from which a given pattern has been extracted by examining the pattern alone. However, the omitted part of the sentence may contain contextual information which specifies the condition when or where the relationship holds. Modeling such contextual information would be useful but is beyond the scope of this work. A consequence of this assumption is that it is possible to batch-annotate a group of inputs that correspond to the same pattern by annotating the pattern itself.
\end{enumerate}

\paragraph{Pattern interpretability} An important consideration in this research is pattern interpretability, which could assist domain experts (who are not NLP experts) in exploring the space of labelling rule antecedents for a given relationship type in a given corpus. Hence, for each pattern, we construct what we call a pattern \textit{lexicalisation}, that is converting a pattern to a readable text-like sequence.

\paragraph{Pattern types} Simple patterns, which can potentially be combined with boolean operators, can be of different types. We illustrate some types of patterns used in our experiments through the following example sentences that include mentions of a gene-disease pair:
\vspace{-0.15cm}
\enumsentence{``We investigate the hypothesis that the knockdown of \textit{BRAF} may affect \textit{melanoma} progression."\label{ex1}} 
\vspace{-0.45cm}
\enumsentence{``The study did not record higher \textit{NF-kb} activity in \textit{cancer} patients."\label{ex2}} 

Below are some types of patterns, as well as their lexicalisations:

\begin{itemize}
    
    \vspace{-0.15cm}\item \textbf{\textsc{keywords}}: words (e.g. `inhibiting') or lemmas (e.g. `inhibit') in the entire sentence or in the text between the entities. This pattern's lexicalisation is, trivially, the word itself. 
    
    \vspace{-0.15cm}\item \textbf{\textsc{path}}: shortest path between the two entity mentions in the dependency graph of the sentence. For instance, in example (\ref{ex1}), the path could be \texttt{\small{\textit{BRAF} <-pobj- of <-prep- knockdown <-nsubj- affect -dobj-> progression -compound-> \textit{melanoma}}}; in example (\ref{ex2}), the path could be \texttt{\small{\textit{NF-kb} <-compound- activity -prep-> in -pobj-> patients -compound-> \textit{cancer}}}). To lexicalise patterns of this type, we extract the nodes (i.e. words) from the path, arrange them as per their order in the sentence and replace the entity mentions by a symbol denoting simply their entity types. For instance, the first pattern becomes \texttt{\small{"knockdown of GENE affect DISEASE progression"}}. This pattern is used extensively in our experiments because it strikes a good balance between expressive power and coverage. We call its lexicalisation a \textbf{simplification} because it is a text-like piece that simplifies a sentence by discarding all but the most essential information.
    
    \vspace{-0.15cm}\item \textbf{\textsc{path\_root}}: the root (word with no incoming edges) of the shortest path between the two entities (e.g. `affect' and `activity' in examples (\ref{ex1}) and (\ref{ex2}) respectively). The lexicalisation could be trivial (i.e. the root itself) or, alternatively, if this pattern is used in an \texttt{\small{AND}} boolean combination with the \textsc{path} pattern, the root can simply be highlighted (e.g. \texttt{\small{"knockdown of GENE \textbf{affect} DISEASE progression"}})
    
    \vspace{-0.15cm}\item \textbf{\textsc{sentence\_root}}: the root of the dependency graph of the entire sentence (e.g. `investigate' and `record' in the examples above), which is often not the same as root of the path connecting the two entities. It can be lexicalised similarly to the pattern above. 
    
    \vspace{-0.15cm}\item \textbf{\textsc{path\_between\_roots}}: the path between the root of the entire sentence and the root of the path between the two entities (e.g. \texttt{\small{investigate -dobj-> hypothesis -acl-> affect}} and \texttt{\small{record -dobj-> activity} for examples (\ref{ex1}}) and (\ref{ex2}) respectively). The pattern can be lexicalised as what we have called ``simplification" (e.g. \texttt{\small{investigate hypothesis affect}}, or, if \texttt{\small{AND}}-ed with the \textsc{path} pattern, all the words from both patterns can be merged and arranged as per their original order in the sentence, potentially with some highlighting to differentiate the two simpler patterns (e.g. \texttt{\small{"\dashuline{investigate} \dashuline{hypothesis} knockdown of GENE affect DISEASE progression"}})
    
    \vspace{-0.15cm}\item \textbf{\textsc{sentence\_root\_descendants}}: the direct descendants of the \textsc{sentence\_root}, for instance, `did', `not' and `activity' in the example (\ref{ex1}), because of the edges \texttt{\small{did <-aux- record}}, \texttt{\small{not <-neg- record}} and \texttt{\small{record -dobj-> activity}}. To lexicalise this pattern, we can extract the words and merge them with words of other patterns. Alternatively, we can devise some simpler sub-patterns, for instance, descendants with \texttt{\small{aux}}, that is auxiliary, edges, such as `may', or descendants with \texttt{\small{neg}} edges such as `not' and place them outside any simplification: \texttt{\small{"\dashuline{investigate} \dashuline{hypothesis} knockdown of GENE affect DISEASE progression \textbf{+} hedging:[may]"}}
    
    \vspace{-0.15cm}\item \textbf{\textsc{path\_root\_descendants}}: the direct descendants of the root of the path between the entities (e.g. `may' and `progression' in example (\ref{ex1}) because \texttt{\small{may <-aux- affect}} and \texttt{\small{affect -dobj-> progression}}; `higher' and `in' in the example (\ref{ex2}) because \texttt{\small{higher <-amod- activity}} and \texttt{\small{activity -prep-> in}}). Its lexicalisation can be the same as that of the previous pattern type.
 
\end{itemize}

Other examples of patterns could be regular expressions or rules informed by an external biomedical ontology (e.g. \texttt{\small{GENE is a Rhodopsin-like receptor}}) or with lexical information from databases like WordNet \cite{wordnet} (e.g. for increasing pattern coverage leveraging synonyms or hypernyms of words in a pattern.)

It should be obvious that the more expressive a pattern becomes (for instance by \texttt{\small{AND}}-ing multiple other patterns), the less capable it is of subsuming many sentences. It is important to discover patterns with this trade-off in mind.

\section{System overview}\label{system-overview}

\begin{figure}[h]
\vspace{-.5cm}
    \centering
    \includegraphics[width=0.9\columnwidth]{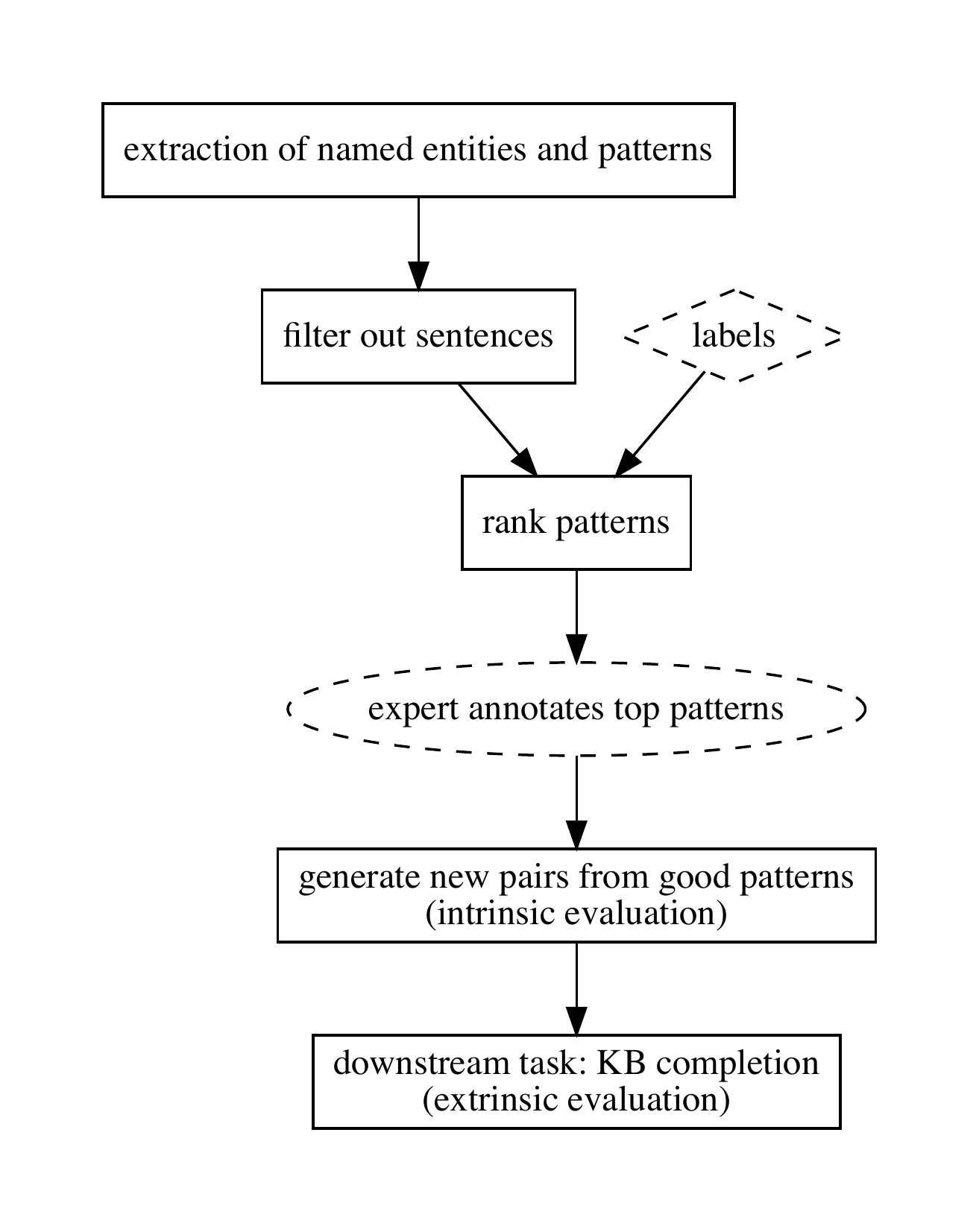}
\vspace{-.8cm}
    \caption{System overview. Diamond box is present only in workflows with seed labels available (i.e. ``no expert but labels" and ``expert with labels"), elliptical box is only present in workflows involving an expert (i.e. ``expert - no labels" and ``expert with labels") and rectangular boxes are always present.}
    \label{fig:workflow}
\end{figure}

In this section, we will describe each step of our system, outlined in Figure \ref{fig:workflow}. 

\subsection{Data preparation} 

\paragraph{Extracting named entities and patterns} The first step is performing named entity recognition (NER) on the sentences in the corpus to enable us to identify all the sentences which contain entity pairs of interest. Our experiments are focused on gene-therapeutic target pairs, however, the system is designed to be agnostic to different types of entities and relationships between them. We then extract the desired patterns from each of these sentences, as described in section \ref{extracting-and-lexicalising-patterns}. For simplicity, we limited our experiments to sentences that contain exactly one gene-disease pair, however, extending the system to handle multiple pairs is straightforward.

We index each sentence in a database along with the lexicalisation for its pattern (e.g. the `simplification' for \textsc{path} or \textsc{path\_between\_roots} patterns) and the entity pair found. This allows us to easily query this database i) for all entity pairs that correspond to a pattern (which is now lexicalised and stored as a string) or ii) for all patterns that correspond to an entity pair.

\paragraph{Filtering out sentences with negation and hedging.} Since we are interested in inputs which unambiguously encode affirmations of facts about entities, we filter out any sentences which contain negation, speculation, or other forms of hedging. We adopt a conservative approach by excluding sentences which match specific instantiations of these pattern types: \textbf{i)} \textsc{keywords} (e.g. presence of terms such as ``no", ``didn't'', ``doubt", ``speculate" etc. in the sentence); our list is modified from NegEx \cite{negex2013}, \textbf{ii)} \textsc{sentence\_root \texttt{\small{AND}} sentence\_root\_descendants} (e.g. ``study we investigated", which makes no statement of results), \textbf{iii)} \textsc{path\_root \texttt{\small{AND}} path\_root\_descendants} (e.g. ``was used", ``was performed"), \textbf{iv)} {path\_between\_roots} (e.g. ``found associated")
This filtering is applied at all stages in our system where sentences are used. 

\subsection{Ranking patterns}\label{workflows}

Below we describe methods for ranking and selecting top patterns in the presence or absence of domain expertise or labelled training data.

\paragraph{Baseline workflow: ``no expert - no labels"} \label{sec:no-expert-no-labels}
In this workflow, we simply extract new pairs using simplifications (from the \textsc{path} pattern type, but other types are also described in our experiments) that have a high enough ($>= 5$) count of entity pairs.

\paragraph{Manual curation in the absence of any labelled training data: ``expert - no labels"} In this workflow, we have a domain expert (a biologist) available for manual curation but there is no labelled training data. It is not possible for a domain expert to annotate all simplifications; this would be too time-consuming. In such cases, active learning can be helpful in deciding which simplifications should be shown to the domain experts for manual curation to best improve the output of our system. The approach that we adopt here is simple but the system could be extended with more sophisticated active learning strategies. We  rank the simplifications by their count of entity pairs; by this we mean the number of unique pairs contained in the sentences in our corpus which correspond to a given simplification (similar to section \ref{sec:no-expert-no-labels}). We then show the top ranked simplifications (i.e. those with the greatest pair count) to our domain expert with a fixed number of random example sentences who then decides if a given simplification is an appropriate heuristic to extract new entity pairs from the corpus, by selecting one of three options ``Yes", ``No", ``Maybe".

\paragraph{Automated workflow: ``no expert but labels"}\label{workflows-automated} For this workflow, a set of gold standard pairs exists as training data but we have no domain experts available. The sentence simplifications can be ranked using various metrics calculated against the gold standard training data. Each simplification is considered as a classifier: A given pair is `classified' by the simplification as positive if the pair can be discovered using the simplification's underlying rule in the corpus. Otherwise, it is classified as negative. The metrics we use to rank the simplifications are precision and recall. The gold standard pairs will form the positive pairs in our training data. To obtain negative pairs, we operate under the closed world assumption: any entity pair found in our corpus of sentences not present in our gold standard set is taken to be negative. This results in an imbalance in the sizes of positive and negative training data which skews the value of precision. To address this, we use a precision metric where the number of true positives and false positives are normalised by the total number of positive and negative pairs respectively in our training data. For each simplification $S$ we define true positives ($TP_S$) and false positives ($FP_S$) as the sets of correctly and incorrectly positive-labelled entity pairs respectively. Our variant of precision for a simplification $S$ is then, $precision_S = \frac{|TP_S| / N_P} {|TP_S| / N_P + |FP_S| / N_N}$ where $N_P$ and $N_N$ are the number of positive and negative pairs respectively in the training data. With this metric, if a simplification classifies 10\% of the positive pairs as true positives and 10\% of the negative pairs as false positives then $precision_S = \frac{0.1} {0.1 + 0.1} = 0.5$. The metric utilises the percentage of each class instead of the absolute number of pairs, as would be the case for the standard precision metric. The definition of recall for a given simplification $S$ is with respect to just the positive training data and is thus unaffected by an imbalance in the sizes.

\paragraph{Manual curation with labelled data: ``expert with labels"} For this workflow, both domain experts and labelled training data are  available to us. We improve on our methodology in the ``expert - no labels" workflow by making use of the metrics discussed in the ``no expert but labels" workflow which are calculated using the labelled training data. As we want to maximise the number and precision of new pairs extracted, we keep only simplifications with recall and precision above certain respective thresholds and present them to domain experts ranked by pair count to ensure they see the most impactful simplifications first.

\subsection{Generating new pairs} \label{sec:find_new_pairs}

All previous stages aim at generating a list of good simplifications. 
We now have a collection of rules which can be used to extract new entity pairs from the corpus. Any simplification selected as useful implies that all entity pairs recovered from the corpus using this rule can be added as positive examples to the dataset. With the selected simplifications, we can batch-annotate thousands of sentences, and hence pairs, with minimal effort. We simply query our database for all new pairs which are found in a sentence expressing any of our selected simplifications.

\paragraph{Clustering simplifications} We found that many simplifications can be very similar up to a few characters. We create clusters of quasi-identical simplifications, and use them i)  to enforce diversity in the selection of simplifications for the user to annotate, by picking only one simplification per cluster and, ii) to safely extend the selection of positive simplification to other simplification in the cluster. We create clusters of simplifications by detecting connected components in a graph where the nodes are the simplifications and the edges are between simplifications which are at a maximum Levenshtein distance of 2. This allows us to be invariant to plural forms, upper/lower case, to short words like in/of etc. Note that some (not all) of these variations could be captured with a lemmatiser. Example of a distance 2 cluster: \newline
\texttt{\small{\{GENE effects on DISEASE}}, \newline\texttt{\small{GENE effect on DISEASE}}, \newline\texttt{\small{GENE effects in DISEASE\}}}\newline
With a distance of 2, we typically increase the number of positive simplifications by 50\%, which significantly increases the recall on new pairs.

\section{Evaluation}\label{evaluation}
We implement two evaluation frameworks. The first is an intrinsic evaluation of the quality of the new extracted pairs. The second is extrinsic; we consider how the inclusion of the new pairs discovered by our system affects the performance of a downstream knowledge base completion task.\footnote{We consider the second type of evaluation extrinsic because knowledge base completion aims to recover latent relationships, whereas knowledge base construction, which the system is built for, is limited to extracting pairs from the literature.}

\subsection{Intrinsic evaluation} \label{sec:intrinsic-eval}

\paragraph{Pair-level} Our aim in this subsection is to construct an intrinsic evaluation framework which can directly measure the quality of the discovered pairs. We do this by holding out a fraction of the gold standard positive pairs and the negative pairs (under the closed world assumption) to be used as a test set. The remaining fraction is used as training data. We evaluate our system by measuring its recall, specificity (true negative rate), precision, and F-score against this test set. In more detail, the new pairs discovered by our selected simplifications are taken to be the positive pairs predicted by our system. The overlap between these new pairs and the positive test set are the true positives ($TP$) while the overlap with the negative test set are the false positives ($FP$). Recall and specificity take their standard definitions. Again, we consider a precision score which is normalised to correct for the imbalance in numbers between positive and negative pairs, $precision = \frac{|TP| / N_P}{|TP|/ N_P + |FP| / N_N}$, where $N_P$ and $N_N$ are the numbers of positive and negative pairs respectively present in the test set (as described in section \ref{workflows-automated}). We take the F-score to be the harmonic mean of this precision variant and recall.

\paragraph{Simplification-level} For the manual workflows, we also consider the expert annotations while assessing the quality of the simplifications. We report $MSP$, the manual simplification precision, based on $N_{Yes}$,  $N_{No}$ and $N_{Maybe}$, the number of simplifications that the expert has annotated as ``Yes", ``No" and ``Maybe". $MSP = \frac{N_{Yes}}{N_{Yes} + N_{No} + N_{Maybe}}$. We expect MSP to be as high as possible.

\paragraph{Extrinsic evaluation via knowledge base completion} The setup for our extrinsic evaluation framework is straightforward and intuitive. The initial gold standard set of positive pairs is split into training and test data. A graph completion model is then trained using the training data and evaluated to determine whether it can predict the existence of the pairs in the test data. To determine whether our knowledge base construction system can add value, we use the new pairs found from our system to augment the training data for the graph completion model, and observe whether this improves its performance against the test set. We use ComplEx \cite{complex}, a well-established tensor factorisation model, as our knowledge base completion model. We provide standard information retrieval metrics to quantify the performance of the graph completion model. These are the precision, $P(k)$, and recall, $R(k)$, calculated for the top $k$ predictions along with the mean average precision ($mAP$). For gene-disease entity pairs, for example, $ mAP = \frac{1}{N_d} \sum \limits_d AveP $, where the sum is over the diseases $d$ with $N_d$ being the total number of diseases, and $AveP = \sum \limits_{k} P(k) \left( R(k) - R(k-1) \right)$ with $P(k)$ and $R(k)$ as defined above.

\section{Main experiments and results}\label{experimental-setup}

\subsection{Datasets}
For all the following experiments, our data was drawn from the following datasets: DisGeNET \cite{disgenet} and Comparative Toxicogenomics Database (CTD) \cite{ctd}. CTD contains two relation types: `marker/mechanism' and `therapeutic'. We use both the entire CTD dataset and the subset of therapeutic gene-disease pairs which we refer to as CTD therapeutic. 

The datasets above are first restricted to human genes and then to the gene-disease pairs which appear in our corpus of sentences; this corpus consists of sentences from PubMed articles which have been restricted, for simplicity, to sentences which contain just one gene-disease pair each. With these restrictions in effect, the CTD dataset has 8828 gene-disease pairs, CTD therapeutic has 169 pairs, and Disgenet has 33844 pairs.

\subsection{Intrinsic evaluation results}

In table \ref{table-main-experiments-intrinsic-eval}, we report the pair-level metrics (see section \ref{sec:intrinsic-eval}) for our three proposed workflows and a baseline (see section \ref{system-overview}). We also report the expert-based metric $MSP$ (see section \ref{sec:intrinsic-eval}) for the two manual workflows. The CTD therapeutic dataset was the most suitable dataset for this evaluation because \textbf{i)} it is very relevant to crucial domains of application such as drug discovery, and \textbf{ii)} its small size makes it a good candidate for expansion. In each session, the expert annotated 200 simplifications accompanied by 20 sentences. It took the expert about 3 hours to annotate the first session, which is a rapid way to generate thousands of new pairs from scratch.

\begin{table*}[h] 
\centering
\small
\begin{tabular}{c|cccccc}
\textbf{Selection method} & \textbf{MSP} & \textbf{New pairs} & \textbf{Recall} & \textbf{Specificity} & \textbf{Precision} & \textbf{F-score} \\ \hline 
expert with labels  & 0.315 & 8875 & 0.286 & 0.976 & 0.923 & 0.436 \\ 
expert - no labels& 0.265 & 9560 & 0.250 & 0.975 & 0.908 & 0.392 \\ 
no expert but labels & - & 30006 & 0.679 & 0.920 & 0.894 & 0.772 \\ 
no expert - no label (baseline)  & - & 59913 & 0.774 & 0.842 & 0.830 & 0.801 \\ \hline 
\end{tabular}
\caption{Intrinsic evaluation results for the main experiments on the CTD therapeutic dataset. This was carried out with a train/valid/test split of 0.4/0.1/0.5, and precision threshold of 0.6 for the `expert with labels' and `no expert but labels' workflows. MSP is our ``manual simplification precision" metric. The precision and F-scores reported here are normalised as described in the section \ref{workflows}.}
 \label{table-main-experiments-intrinsic-eval}
\end{table*}

We find that our three main proposed workflows (`expert - with labels', `expert - no labels', and the fully automated `no expert but labels') all discover a significant number of new gene-disease therapeutic pairs. As confirmed by both pair-level and user-based metrics, incorporating the use of domain expert's time and the use of labelled data results in higher precision at the expense of recall.

\subsection{Extrinsic evaluation results}

In table \ref{table-main-experiments-extrinsic-eval}, we list the results of the downstream knowledge base completion task for the fully automated workflow and the baseline. We compare the performance of our knowledge base completion model when trained with just the initial seed training data versus the seed training data augmented with the new pairs discovered by our fully automated workflow (and baseline workflow). 

The addition of new pairs from the fully automated workflow gives us a higher mean average precision ($mAP$) than with just the seed dataset. We obtain a higher precision (for the top 100 and top 1000 predictions) while maintaining the same level of recall. For the baseline workflow, $mAP$ is higher but with lower precision (for the top 100 and 1000 predictions respectively).

\begin{table*}[h]
\centering
\small
\begin{tabular}{c|cccc}
\textbf{Selection method} & \textbf{New pairs} & \textbf{MAP} & \textbf{Precision} & \textbf{Recall} \\ \hline 
seed dataset only  & - & 0.0414 & 0.0179 / 0.0179 & 1.0 / 1.0 \\ 
no expert but labels & 30006 & 0.0545 & 0.0192 / 0.0192 & 1.0 / 1.0 \\ 
no expert - no label (baseline)  & 59913 & 0.1885 & 0.01 / 0.0015 & 0.6019 / 0.9208 \\  \hline 
\end{tabular}

\caption{Extrinsic evaluation results for the CTD therapeutic dataset. The experiment parameters are the same as those given in table \ref{table-main-experiments-intrinsic-eval}. Precision figures are given as `top 100 / top 1000' and similarly for recall.} 
\label{table-main-experiments-extrinsic-eval}
\end{table*}

\section{Top simplifications} \label{top-simplifications}
In table \ref{table:top-simplification}, we show the simplifications with the highest count of Disease-Gene pairs in our whole corpus (after the sentence filtering), which have been annotated by the expert as ``Yes" or ``No", for the CTD therapeutic dataset. While ``Yes" and ``No" patterns look similar, we can clearly see differences in language. The ``No" annotations look unspecific while the ``Yes" ones express the target has a therapeutic effect on the disease.  

\begin{table}[h]
    \begin{minipage}[t]{.55\linewidth}
      \tiny
        \begin{tabulary}{\columnwidth}{@{\hskip0pt}R@{\hskip-2pt}R@{\hskip3pt}|} 
        \small{\textbf{Pairs}} & \small{\textbf{``Yes" simplif.}} \\ \hline 
        3345  &  \texttt{role of GENE in DISEASE}\\
        839  &  \texttt{GENE plays in DISEASE}\\
        648  &  \texttt{GENE involved in DISEASE}\\
        321  &  \texttt{GENE target in DISEASE}\\
        318  &  \texttt{GENE target for DISEASE}\\
        289  &  \texttt{GENE mice develop DISEASE}\\
        279  &  \texttt{DISEASE caused by mutations in GENE}\\
        276  &  \texttt{GENE gene for DISEASE}\\
        273  &  \texttt{role of GENE in development of DISEASE}\\
        237  &  \texttt{GENE promotes DISEASE}  \\
          &  
        \end{tabulary}
        
    \end{minipage}%
    \begin{minipage}[t]{0.45\linewidth}
        \tiny
        \begin{tabulary}{\linewidth}{@{\hskip1pt}R@{\hskip-1pt}R@{\hskip2pt}} 
        \small{\textbf{Pairs}} & \small{\textbf{``No" simplif.}} \\ \hline
6629  & \texttt{GENE DISEASE}\\
4110  & \texttt{DISEASE GENE}\\
3350  & \texttt{GENE and DISEASE}\\
2370  & \texttt{GENE in DISEASE}\\
2333  & \texttt{DISEASE and GENE}\\
1228  & \texttt{GENE DISEASE cells}\\
904  & \texttt{DISEASE of GENE}\\
879  & \texttt{DISEASE in GENE}\\
638  & \texttt{DISEASE in GENE mice}\\
572  & \texttt{role for GENE in DISEASE}\\
528  & \texttt{GENE in DISEASE patients}
\end{tabulary}
    \end{minipage} 
    \vspace{-0.3cm}
\caption{Top 10 simplifications for CTD Therapeutic annotated ``Yes'' (left) and ``No" (right) by the expert.}
\label{table:top-simplification}
\end{table}

\section{Further experiments} \label{further-experiments}

We performed several other experiments using our fully automated workflow to evaluate the quality of the new pairs discovered as we varied our experiment parameters.

We consider three dimensions of variation: varying the precision threshold for selecting simplifications, varying the size of the seed training set, and varying the expressiveness of the simplification (for example, by including the \textsc{sentence\_root} or restricting to simplifications with at least a specified number of words).

The intrinsic evaluation results for these experiments are listed in tables \ref{table-intrinsic-vary-thresholds}, \ref{table-intrinsic-splits}, and \ref{table-intrinsic-simplification-type}.
In all cases, as we make our system more selective either by raising the precision threshold, by starting with fewer seeds pairs, or by restricting to more informative simplifications, we unsurprisingly obtain higher precision at the expense of lower recall. 

The extrinsic evaluation framework is less sensitive to these changes but improvements were observed (without any noticeable trend) for all these parameter changes. 

\begin{table}[h]
\centering
\small
\begin{tabulary}{\columnwidth}{C C|C C C C C}
\textbf{Dataset} & \textbf{Thres.} & \textbf{New pairs} & \textbf{R} & \textbf{S} & \textbf{P} & \textbf{F} \\\hline
CTD & 0.8 & 29592 & 0.297 & 0.918 & 0.783 & 0.430 \\ 
CTD & 0.4 & 50329 & 0.379 & 0.863 & 0.735 & 0.500 \\ 
\hline
DG & 0.8 & 17441 & 0.180 & 0.947 & 0.773 & 0.292 \\ 
DG & 0.4 & 45446 & 0.314 & 0.867 & 0.703 & 0.434 \\
  \hline 
\end{tabulary}

\caption{Intrinsic evaluation results (Recall, Specificity, Precision and F-score) on CTD and DisGeNET (DG) as we vary the precision threshold for the `no expert but labels' workflow. Experiments are done with a train/valid/test split of 0.8/0.1/0.1 and we restrict to simplifications with at least 5 words to ensure that they are reasonably expressive.}
\label{table-intrinsic-vary-thresholds}
\end{table}

\begin{table}[h]
\centering
\small
\begin{tabulary}{\columnwidth}{C|C C C C C}
\textbf{Train/val/test} & \textbf{New pairs  } & \textbf{R} & \textbf{S} & \textbf{P} & \textbf{F} \\\hline 
0.8/0.1/0.1 & 29592 & 0.297 & 0.918 & 0.783 & 0.430 \\ 
0.5/0.1/0.4 & 25539 & 0.274 & 0.930 & 0.798 & 0.408 \\  
0.2/0.1/0.7 & 18268 & 0.225 & 0.950 & 0.818 & 0.352 \\\hline 
\end{tabulary}
\caption{Intrinsic evaluation results (Recall, Specificity, Precision and F-score) for CTD as we vary size of the seed training data for the `no expert but labels' workflow. Experiments are done with a precision threshold of 0.8 and we restrict to simplifications with at least 5 words to ensure that they are reasonably specific.}
\label{table-intrinsic-splits}
\end{table}

\begin{table}[h]
\centering
\small
\begin{tabulary}{\linewidth}{@{\hskip-2pt}C@{\hskip-1pt}C|CCCCCC}
\textbf{Min length} & \textbf{\textsc{\small{sentence root}}} & \textbf{New pairs} & \textbf{R} & \textbf{S} & \textbf{P} & \textbf{F} \\ \hline 

5 & No & 29592 & 0.297 & 0.918 & 0.783 & 0.430 \\                 
5 & Yes & 13284 & 0.174 & 0.963 & 0.824 & 0.288 \\                
7 & No & 5313 & 0.066 & 0.986 & 0.824 & 0.122  \\
7 & Yes & 1814 & 0.028 & 0.995 & 0.856 & 0.055 \\ \hline 
\end{tabulary}
\caption{Intrinsic evaluation results (Recall, Specificity, Precision and F-score) for CTD as we vary simplification expressive power (minimum length) for the `no expert but labels' workflow. Experiments are done with a train/valid/test split of 0.8/0.1/0.1 and a precision threshold of 0.8.}
\label{table-intrinsic-simplification-type}
\end{table}

\section{Conclusions and further work}\label{conclusions-and-further-work}

We have presented a simple and effective method for knowledge base construction when the desired relational data are scarce or absent. We have demonstrated its effectiveness via \textbf{i)} classification metrics on a held-out test set, \textbf{ii)} human evaluation and \textbf{iii)} performance on a downstream knowledge base completion task. We further show that in the presence of a small set of data, it is possible to control the quality of the pairs discovered, by introducing stricter precision thresholds when ranking patterns. Our method could in principle be extended in order to: \textbf{1)} handle higher-order (e.g. ternary) relations between tuples, as opposed to pairs (for instance using dependency subgraphs that connect more than two entities cooccurring in a sentence), \textbf{2)} discover explicit negative examples of a binary relation instead of simply positive examples, \textbf{3)} train sentence-level relation extraction systems, \textbf{4)} collect and utilise continuous, rather than discrete annotations for each pattern (e.g. annotators could indicate the percentage of correct example sentences that correspond to a pattern displayed) as part of a more sophisticated active learning strategy, \textbf{5)} extract patterns from a semantic representation \cite{banarescu-EtAl:2013:LAW7-ID} and, finally, \textbf{6)} map patterns to a vector space using a distributional representation (e.g. defined by their neighbouring words in sentences) and cluster them for an optimal balance between expressive power and coverage.

\section*{Acknowledgments}

We are very grateful to Nathan Patel for his engineering support and to Alex de Giorgio for his thorough feedback and domain expertise. We would also like to thank many of our colleagues working on drug discovery and link prediction for insightful conversations, as well as Felix Kruger for proofreading the final version of this paper.

\nocite{*} 

\end{document}